\documentclass[11pt,a4paper]{article}
\usepackage[hyperref]{acl2021}
\usepackage{times}
\usepackage{latexsym}
\usepackage{graphicx}
\usepackage{CJKutf8}
\usepackage{verbatim}
\usepackage{enumitem}

\usepackage{microtype}

\aclfinalcopy 

\title{\textsuperscript{Meta}SRL++: A Uniform Scheme for Modelling Deeper Semantics}

\author{
    Fritz Hohl\thanks{Sony Europe B.V.}, 
    Nianheng Wu\thanks{Sony Europe B.V.}, 
    Martina Galetti\thanks{Sony Computer Science Laboratories Paris}, 
    Remi van Trijp\thanks{Sony Computer Science Laboratories Paris}\\
    \texttt{fritz.hohl@sony.com}}

\date{}

\begin{document}
\maketitle
\begin{abstract}
Despite enormous progress in Natural Language Processing (NLP), our field is still lacking a common deep semantic representation scheme. As a result, the problem of meaning and understanding is typically sidestepped through more simple, approximative methods. This paper argues that in order to arrive at such a scheme, we also need a common \textit{modelling} scheme. It therefore introduces \textsuperscript{Meta}SRL++, a uniform, language- and modality-independent modelling scheme based on Semantic Graphs, as a step towards a common representation scheme; as well as a method for defining the concepts and entities that are used in these graphs. Our output is twofold. First, we illustrate \textsuperscript{Meta}SRL++ through concrete examples. Secondly, we discuss how it relates to existing work in the field. 

\end{abstract}

\section{Introduction}

A quick glance at the available NLP tools and corpora quickly reveals that there are much more resources available for syntactic than for semantic analysis. Therefore, if a particular application requires a deep \textit{understanding} of its input, the requirement of semantics is typically sidestepped. Instead, meaning and understanding are either approximated through less complicated mechanisms, or left up to the obscure inner workings of neural approaches (often paying the price of needing a larger labelled corpus or needing extensive computing resources such as GPU clusters).

We argue that one reason for the lack of resources for deep semantic analysis is due to the lack of a common, uniform representation scheme that is able to represent \textit{all} aspects of semantics. That is not to say that there is not already a lot of extensive research performed in different areas of semantics: it is rather that these efforts need to develop their own specific-problem-related notations because of this lack. 

It is our belief that in order to arrive at a common representation scheme, we also need to develop a common \textit{modelling} scheme that allows to model semantics. By ``modelling scheme'' we mean formalisms similar to the Unified Modeling Language (UML) for software Engineering. UML provides a mechanism for developers to model software systems by letting them identify components and relations between these components. The same mechanism can be used to model a system at different detailing levels thus allowing for a hierarchical model.

This paper aims to contribute to the goal of a common semantic representation scheme by presenting \textsuperscript{Meta}SRL++, a uniform \textit{modelling} scheme for all types of semantic information using Semantic Graphs. The paper is structured as follows: Section \ref{s:semantic-graphs} introduces Semantic Graphs, which are diagrams that model some semantic content. Section \ref{s:muhai} then demonstrates the usage of \textsuperscript{Meta}SRL++ on semantic data from the European Pathfinder project \href{https://muhai.org/}{MUHAI}. We then discuss related work (Section \ref{s:related-work}) and conclude the paper with a potential extension of our modelling scheme called SRL++, a semantic representation scheme based on \textsuperscript{Meta}SRL++.

\section{Semantic Graphs}
\label{s:semantic-graphs}

As its name implies, \textsuperscript{Meta}SRL++ subscribes to the longstanding history in cognitive science to operationalize semantic information as frames \cite{minsky_framework_1975,fillmore_frame_1976} or schemas \cite{rumelhart_schemata_1980} that capture the recurrent aspects of experience, which is also pursued in various Semantic Role Labelling (SRL) approaches such as FrameNet \cite{baker_berkeley_1998} and PropBank \cite{palmer2005propbank}. That is, it describes the semantic elements of a document as well as the roles that these semantic elements play with respect to other elements.

In contrast to existing SRL approaches \cite[e.g. PropBank;][]{palmer2005propbank}, all elements in our semantic modelling scheme are semantic units (and not simply parts of a textual sentence), and predicates are not only (or, at least, mainly) derived from verbs. Instead, predicates can represent all sorts of semantic information. Therefore, they could have been derived from all sort of combination of information in texts or even other modalities. As we decided to model only semantic information, there is, in principle, no association of the semantic information to the document it was created from (out of practical reasons this association can be created, though, see Section \ref{s:causation-in-italian}).

Our modelling scheme is realized by \textit{Semantic Graphs}. Semantic Graphs consists of different elements that fall into two categories: nodes and labelled edges.

\subsection{Nodes}
\label{s:nodes}

There are three kinds of nodes in our Semantic Graphs: \textbf{concepts}, \textbf{entities} and \textbf{ommitted nodes}.

\textit{Concept} nodes (representing predicates) are the main building block of Semantic Graphs. They each represent a single semantic aspect. They are depicted as a box with the concept name in it. \textsuperscript{Meta}SRL++ does not dictate the set of concepts to choose from (in the same way that UML does not dictate the content of e.g., a component). It is the responsibility of the author of a Semantic Graph to specify concepts in a way that the readers can understand a graph.

Concepts are however not enough to describe all the semantics we need to model: we also need \textit{entities}. Entities are individual instances of one or more classes, i.e. in order to understand them it is not only important to know their distinguishing feature (e.g. a name or a value), but also the classes they are an instance of. The most notable examples are Named Entities, but also other objects are typically given entity status.

Fortunately, entity semantics is the one field of semantics where there is already a commonly used notation and where there are sufficient tools to handle them, so \textsuperscript{Meta}SRL++ does not aim to reinvent the wheel: it adopts the standard definitions of concepts as generalizations of entities of the same type, and treats entity classes as concepts.

In contrast to concepts, entities are always leaves in the Semantic Graph; they do not have edges leading away from them. We depict them as circles with a value label. Optionally, there can be one or more concept names on top of the value label. These concepts are some of the classes this entity belongs to.

Finally, we also include leaf nodes that are called ``ommitted nodes'', depicted as a grey circle. These nodes are used for capturing semantic elements that are implied by a sentence, but which is missing in the surface text (\textit{Null Instantiations} as FrameNet calls them; also known as \textit{Implicit Arguments}).

\subsection{Labelled Edges}
\label{s:labelled-edges}

Concepts are connected to other nodes via directed Labeled Edges (although the arrows are often omitted if the direction is obvious). These edges represent role relations of the connected nodes to the concept. Therefore, the edge labels are always one of the role names from the concept from which they are originating. They are depicted as directed lines that originate at the bottom of the concept they relate to, and lead to the top of nodes that take the roles regarding this concept.

There is a special role that can be used: the indexed role. This role is indexed by a positive Integer starting from 1. It models the case that sometimes there is a multitude of elements in the same role, e.g. child roles of a ``parent'' concept. In order to prevent the necessity to define a separate role for any (finite) number of such children in a parent concept, the use of an indexed role reduces definition overhead. An example of an indexed role can be found in Fig. \ref{fig:semantic-graph-in-MetaSRL++}.

\begin{figure}[t] 
\centering
\includegraphics[trim={0 0.4cm 0 0.3cm},width=\columnwidth]{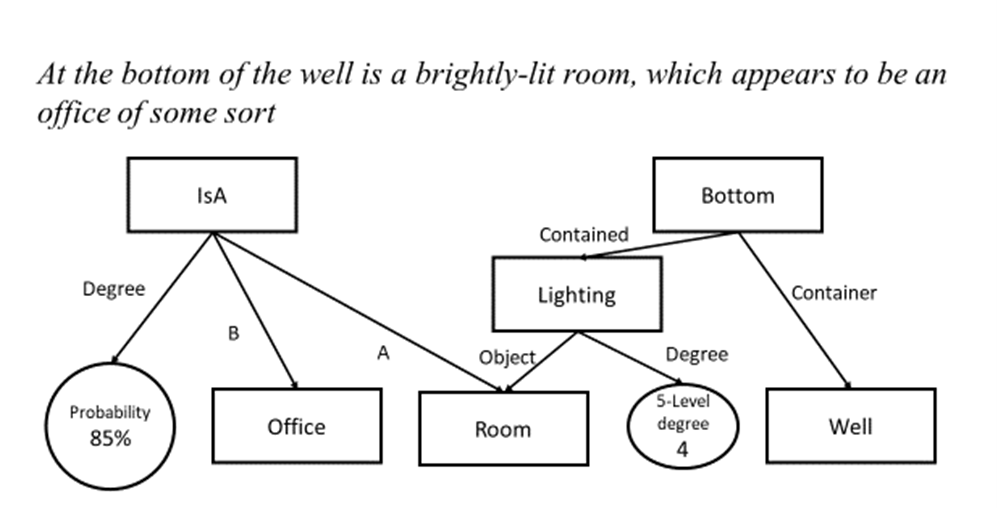}
\caption{Example of a Semantic Graph for the sentence ``at the bottom of the well is a brightly-lit room, which appears to be an office of some sort''}
\label{fig:example-semantic-graph}
\end{figure}

\subsection{An example}
\label{s:semantic-graph-example}

Let us now look at an example in Figure \ref{fig:example-semantic-graph}, which shows a sentence (``at the bottom of the well is a brightly-lit-room which appears to be an office of some sort'') and its manually-constructed semantic graph. In this sentence, there are two parts which can be divided into two subgraphs. 

The subgraph on the right represents a room, located at the bottom of a well, that is lit to a rather high degree. The top-most node is the  {\em Bottom} concept. If we look up our concept catalogue, we will find that this concept has two roles, a ``Container'', and a ``Contained''. In the description of the {\em Bottom}concept we might find that the ``Container'' role represents the outlining element, and the ``Contained'' the element that sits at the bottom of the ``Container''. Therefore, the ``Container'' role is given to the  {\em Well} concept and the ``Contained'' to the lighted room. Equally, the  {\em Lighting} concept might have two roles, the lit ``Object'' and the ``Degree'' to which the ``Object'' is lit. The Object is obviously the {\em Room}. The ``Degree'' is filled by an entity node. This entity node has the value {\em 4} and is of a concept {\em 5-level degree}. This construction results from mapping a textual expression to a semantic representation of this expression. This has the advantage of maybe covering multiple textual expressions to the same representation (e.g. ``very bright''), to be independent of the semantics of the English expression (``brightly-lit'') which might not have an equivalence in other languages, and to be much more understandable digitally. Importantly, the {\em Lighting} concept might very well have more possible roles (e.g. a light source), but not all of them are filled in this sentence.

In the same manner the left side of the Semantic Graph can be interpreted. In fact, in the subgraph on the left, the {\em IsA} concept has two roles with the meaning “A is equal to B”. We modeled the expressions “appears” with a probability entity in the role ``Degree'' of the {\em IsA} concept.

The two sides of the Graph are connected by sharing the {\em Room} concept. This could have been also the results of two sentences (“At the bottom of the well is a brightly-lit room. This room appears to be an office of some sort”). It is also conceivable to take the right subgraph and put it completely (i.e. starting at {\em Bottom}) in the role A position of {\em IsA}, but this would represent more the sentence “At the bottom of the well a brightly-lit room appears to be an office of some sort”).

This Semantic Graph is a representation of the semantics of the example sentence that abstracts away from actual linguistic expressions.\footnote{The graph doesn't show the narrative time concept, which means the default present time is assumed.} This means that the same model can also be used for other languages such as Chinese, where the sentence would appear as ``\begin{CJK*}{UTF8}{gbsn}
在井底有一间光亮的小室，可能是一间办公室
\end{CJK*}''. Moreover, the representation is modality-independent, and could be used for modelling e.g. a movie scene. The scene might start with the brightly-lit room with fuzzy focus, and then gradually clear the focus to reveal the details of an office.

As with any modelling scheme, some choices remain up to the discretion of the modeller. It may, for instance, not always be clear whether to model a concept as an entity or vice versa. As a rule, entities cannot have outgoing edges (i.e. roles). If edges are needed, you need to use the concept form. Only leaf nodes can be modeled as either concepts or entities.

One of the consequences of semantic parsing is that the same word can be parsed into completely different concepts. For example, the word “it” might be modelled as a concept that refers to a reference of a single, 3rd person (in semantic, not syntactic terms) entity, an entity of the Movie concept; or it can be part of concepts that model multi-word expressions (like in “Hold it!”).

\section{Usage of MetaSRL++ in MUHAI}
\label{s:muhai}

\begin{figure}[t]
  \includegraphics[width=\columnwidth]{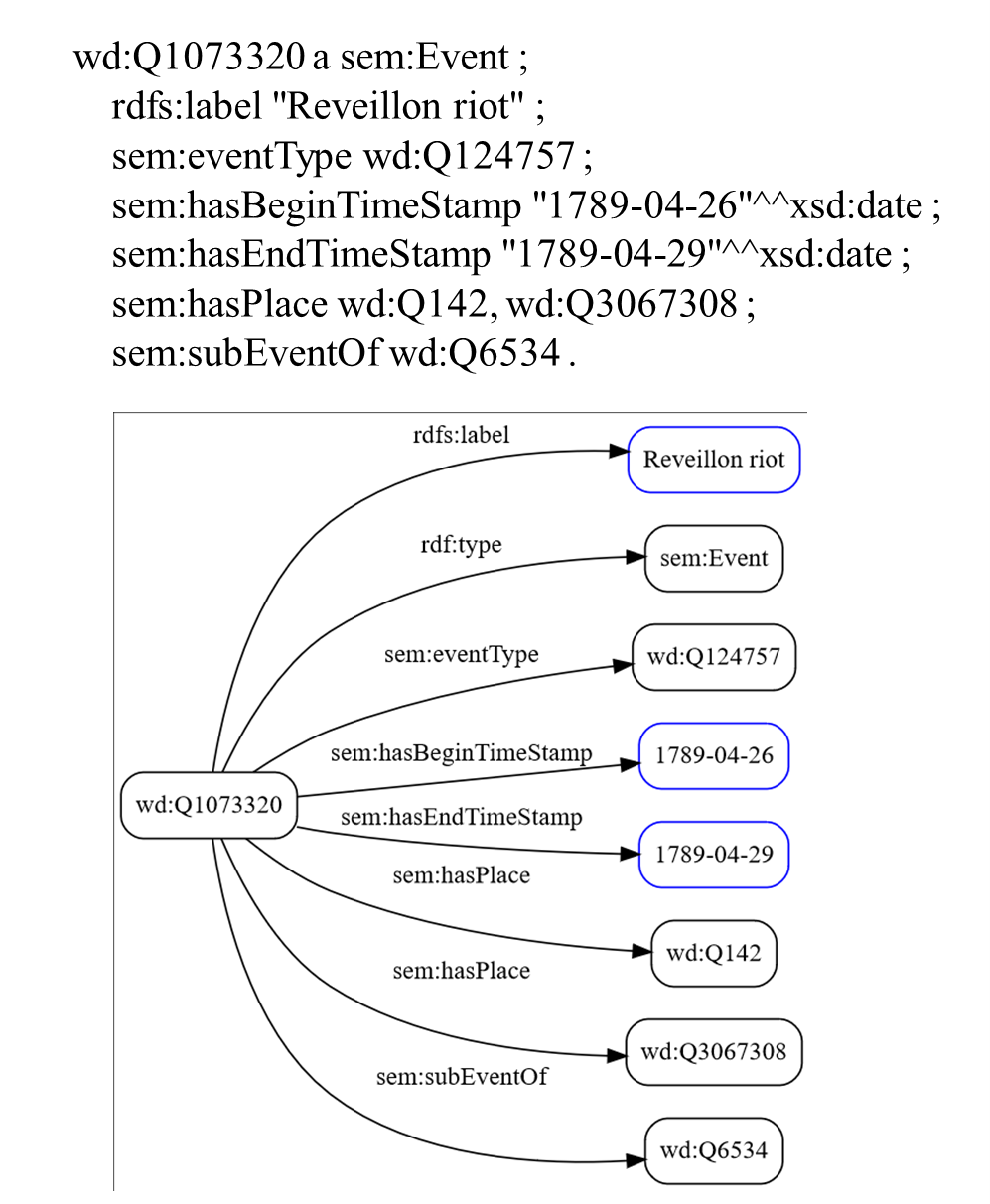}
  \caption{Example Raw Data from a Knowledge Graph.}
  \label{fig:example-raw-data}
\end{figure}

\href{https://muhai.org/}{MUHAI (Meaning and Understanding in Human-centric AI)} is a European Pathfinder project that studies how to develop \textit{meaningul} AI. `Meaningful' here means AI systems that complement the reactive behavior of current-generation AI systems with rich models of problem situations in domains for more deliberate reasoning. The MUHAI project includes a diverse set of case studies, ranging from everyday activities such as cooking to social media observatories and artwork interpretation \cite{steels22muhai}. 

\begin{figure*}[t]
\centering{
  \includegraphics[trim={0 0 0 3cm},clip,width=0.8\textwidth]{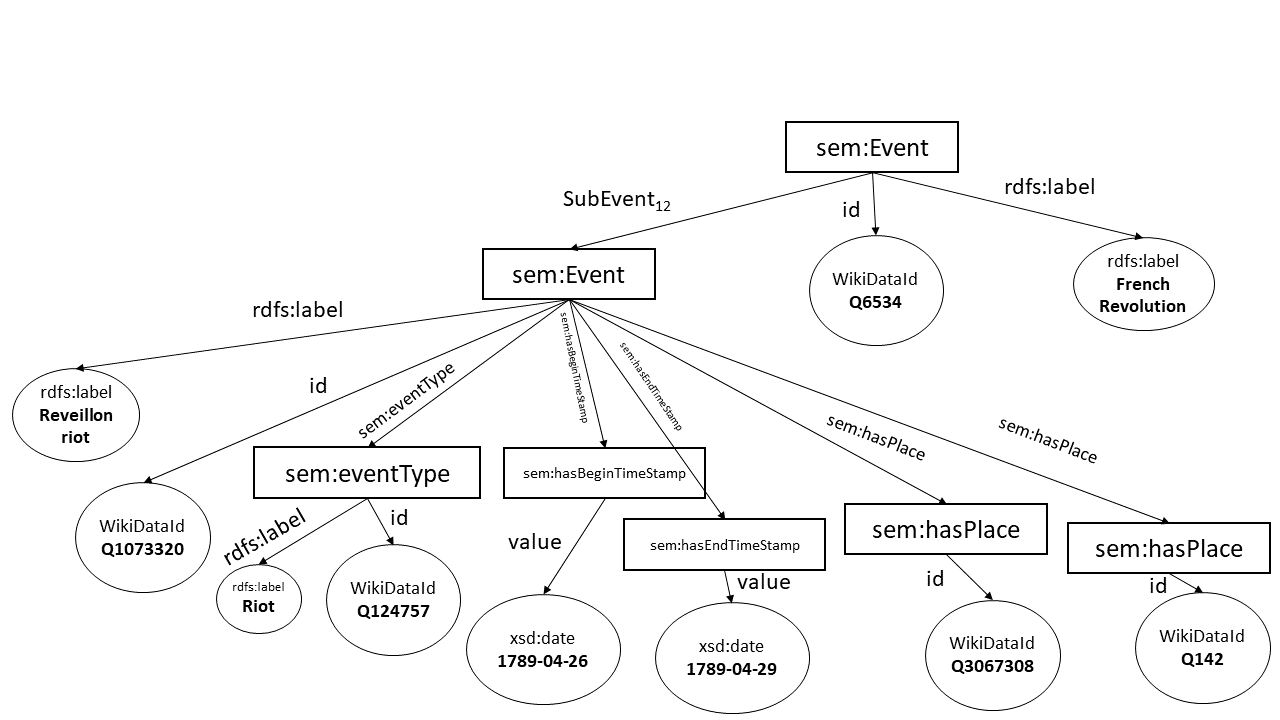}
  \caption{Semantic Graph in \textsuperscript{Meta}SRL++}
  \label{fig:semantic-graph-in-MetaSRL++}}
\end{figure*}

These different subprojects produce different types of semantic data in different natural languages. In order to export this data into a suited, uniform format and to offer applications of this data a uniform format, MUHAI selected \textsuperscript{Meta}SRL++ as its overall modelling scheme. To that end, a Python library was created that allows to read and write the \textsuperscript{Meta}SRL++ XML format.

\subsection{Historic Events from Knowledge Graphs}

As a first example, please find in Fig. \ref{fig:example-raw-data} a piece of raw semantic data, in this case some parts of a knowledge graph containing information on historic events of the French Revolution \cite{blin22narrative} in form of parts of a Turtle (.ttl) file. The upper part of Fig. \ref{fig:example-raw-data} contains the file content, the lower part a graphical representation of the same content.

Using this example, we modelled this data in \textsuperscript{Meta}SRL++ (see Fig. \ref{fig:semantic-graph-in-MetaSRL++}). Knowledge graphs represent information using semantic triples (subject, predicate, object). Since entities cannot have outgoing edges in \textsuperscript{Meta}SRL++, all predicates that link two entities in the knowledge graph were elevated to the status of concept nodes, while the entities simply remain entities. In this case it is not really possible to have human-friendly labels for the entities, so we simply used the same label for the edges. Edges to entities were labelled with “id” if they lead to entity labels of the knowledge graph and with ``value'' in all other cases. This choice of edge labelling implies that the corresponding concepts contain theses labels as possible roles.

There are two differences to this rather mechanic way of translating these Knowledge Base triples. First, we decided that the top-level concept of this type of data is an event of the type “sem:Event” (as the entry states anyway), and that the main entity (in this example wd:Q1073320) should be added as an “id” role of this concept. Also, the “rdfs:label” relation was modelled as the corresponding role of that concept. The second difference is the handling of the “sem:subEventOf” relation. In order to model the sub events of an event more explicitly (as events are the main content of this data), also sub events were modelled as roles of an encompassing event. In this case this role was modelled as an indexed role (the “12” in Fig. \ref{fig:semantic-graph-in-MetaSRL++} serves only illustratory purposes to show that the top event could have a whole number of sub events).

For the above method, obviously, the semantics of the concepts are defined by the definition of the relations. As the latter are quite well defined, also the first can be understood. Roles, obviously, play only a small role in this example.

\subsection{Causation in Italian Sentences}
\label{s:causation-in-italian}

\begin{figure}[h]
  \includegraphics[trim={0 0.1cm 0 0},clip,width=\columnwidth]{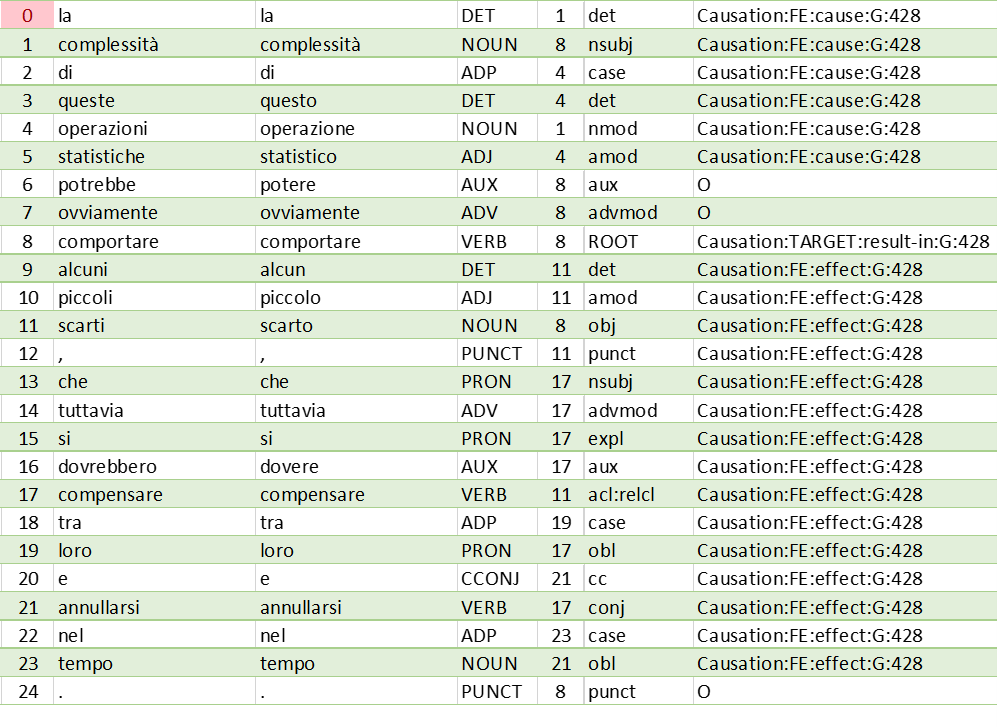}
  \caption{Example Sentence in Italian}
  \label{fig:example-sentence-italian}
\end{figure}

In a second example we have Italian sentences in a CoNLL-style format with additional annotations about cause and effect relations (see Figure \ref{fig:example-sentence-italian}), inspired by earlier work on causal semantic frames in English \cite{BeulsVanEeckeCangalovic+2021}. A semantic frame (such as the Causation frame) can be straightforwardly modeled as a concept; and its frame elements (e.g. cause, effect, and so on) as its roles.

\begin{figure*}[h]
  \includegraphics[trim={0 0 0 2.5cm},clip,width=\textwidth]{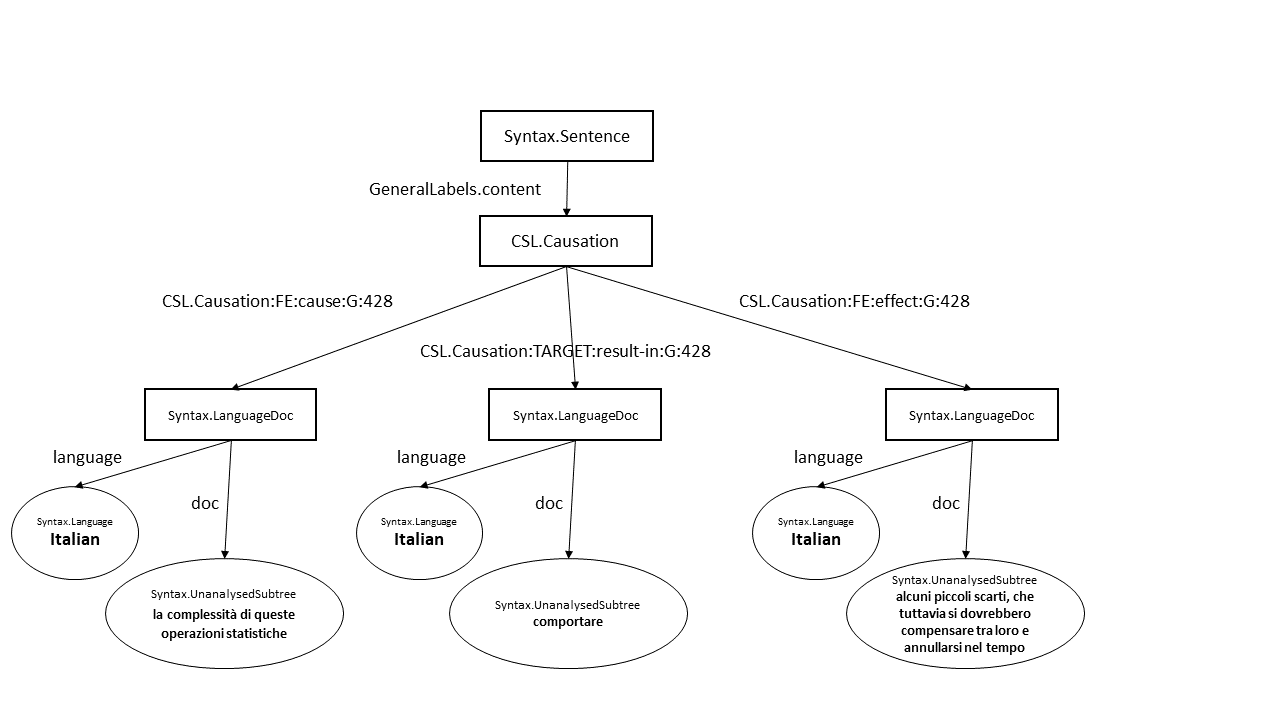}
  \caption{Semantic Graph}
  \label{fig:semantic-graph-italian}
\end{figure*}

Interesting to note is that the original data only provided the causation labels as semantic data, while all other fields contain standard syntax-related information such as lemmas, POS tags, and (dependency) parsing labels, as well as the tokens of the original sentence. We only modelled the sequences of the original sentence that belong to the corresponding causation elements (which were, incidentally or not) also subtrees of the parse tree). 

From our point of view, such sequences have normally no business in a pure Semantic Graph as they are language-dependent and contain semantic information only indirectly as natural language. However, for some use cases it may be more efficient to keep the connection between semantics and surface texts; or sometimes we need to be able to model the relation between nonsensical phrases and otherwise purely semantic content (as in ``and then she said `Hnnngom' or something which I did not understand.''). To illustrate such use cases, we explicitly modelled surface sequences as entities of the concept UnanalysedSubtree, which are elements of a LanguageDoc which also has a language role. Finally, we also wanted to keep the information that the text portion that was analysed was a sentence. We are of the opinion that sentences are primarily non-semantic entities (because the question of how to portion semantic content into sentences is more a cultural aspect that can be answered differently for text generation depending on e.g. the expected literary abilities of the target audience). Therefore, we put a Sentence concept at the top.

\section{Related Work}
\label{s:related-work}

Our scheme aims to be a \textit{modelling} scheme, i.e. a kind of meta representation for semantics (that also claims to be able to model a large amount of semantic aspects). To our knowledge, no other meta representation scheme exists as all of the related work are approaches that are targeted towards representing semantics concretely. Therefore, we decided to relate our scheme to other work by roughly outlining how these semantic representation schemes can be modelled using our approach. Out of space restrictions, we have to restrict ourselves to three semantic representation schemes: AMR, UMR, and UCCA.

\subsection{Abstract Meaning Representation (AMR)}
\label{s:AMR}

{\href{http://amr.isi.edu/}{AMR}} is a notation based on PENMAN. From a structural point of view, AMR consists of nodes which are labelled each with a variable name and a concept label and labelled edges (which represent relations)~\cite{banarescu2013abstract}. The semantic concepts are the nodes of the graph and the edges represent the relations that bound the different nodes. Every semantic concept and every node in the graph is assigned to a variable and it is labeled with English words (ex: boy ?b), PropBank notation (ex: say-01 ?s) or, in certain cases, by special keywords (ex: amr-unknown ?a). The possible relations between the edges can be represented by Frame Arguments (ex: :arg0), general semantic relations (ex: :polarity), relations for quantities (ex: :quant), for time (ex: :time) and for lists (ex: :op1). More in detail, every AMR graph has a unique root, displayed as the top node in the tree, variables (e, x, y, etc.), events, concepts (ex: boy) and roles (ex : ARG0, ARG1, etc.). A property of AMR graphs is their ability to invert roles (the relations are semantically equivalent, but are structured differently). It must be underlined that Abstract Meaning Representation is geared towards English and the vocabulary of English~\cite{xue2014not}, even if some efforts had been made to apply it to other languages (parser in Chinese, French, German, Spanish, Japanese)~\cite{vanderwende2015amr}. 

This structure can be converted to \textsuperscript{Meta}SRL++ by replacing:

\begin{itemize}[noitemsep,topsep=3pt,parsep=0pt,partopsep=0pt]
    \item nodes by concepts
    \item constants by entities
    \item labelled edges by labelled edges
    \item the use of a variable reference by connecting an edge to the corresponding node which, in AMR, had the corresponding variable name associated with (if the graphical form of AMR is used, this is already done)
\end{itemize}

and by moving all outgoing edges of nodes to the bottom and all ingoing ones to the top and by removing all variables.

\subsubsection{Example: from AMR to \textsuperscript{Meta}SRL++ }

\begin{figure}[h]
  \includegraphics[width=0.8\columnwidth]{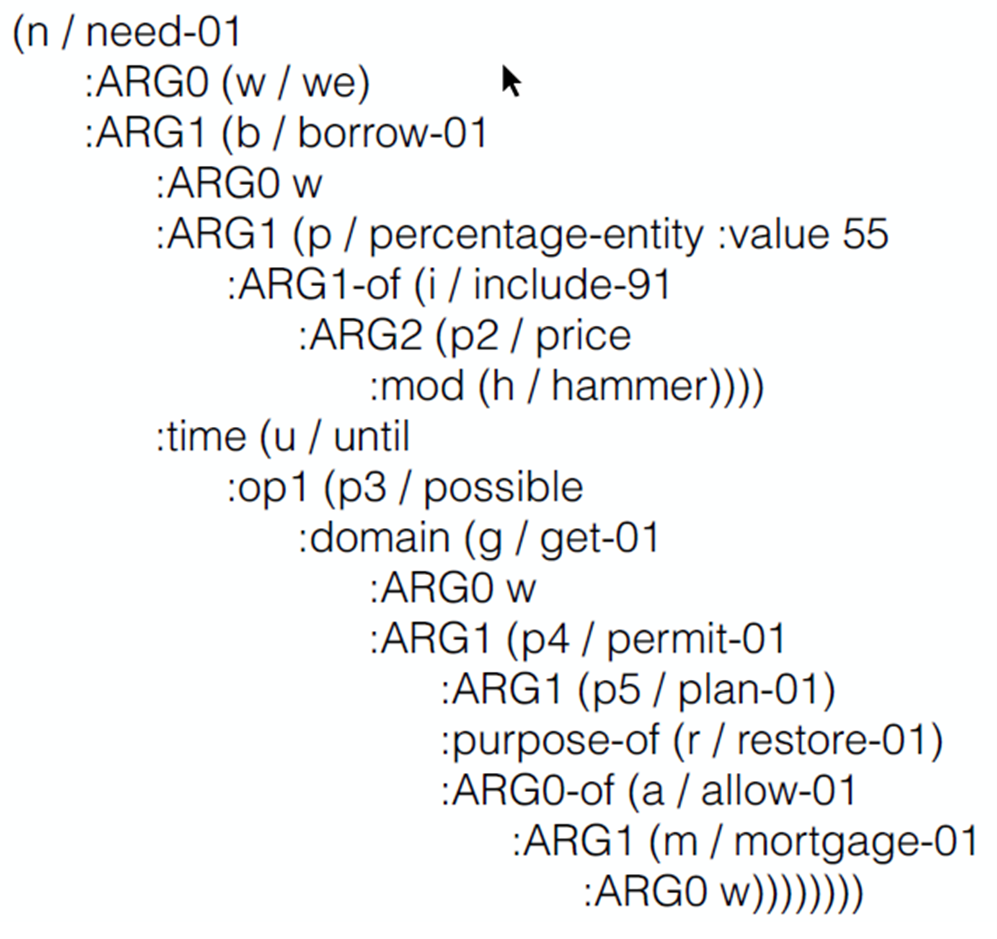}
  \caption{Example Sentence as AMR}
  \label{fig:example-sentence-AMR}
\end{figure}

\begin{figure}[h]
  \includegraphics[width=\columnwidth]{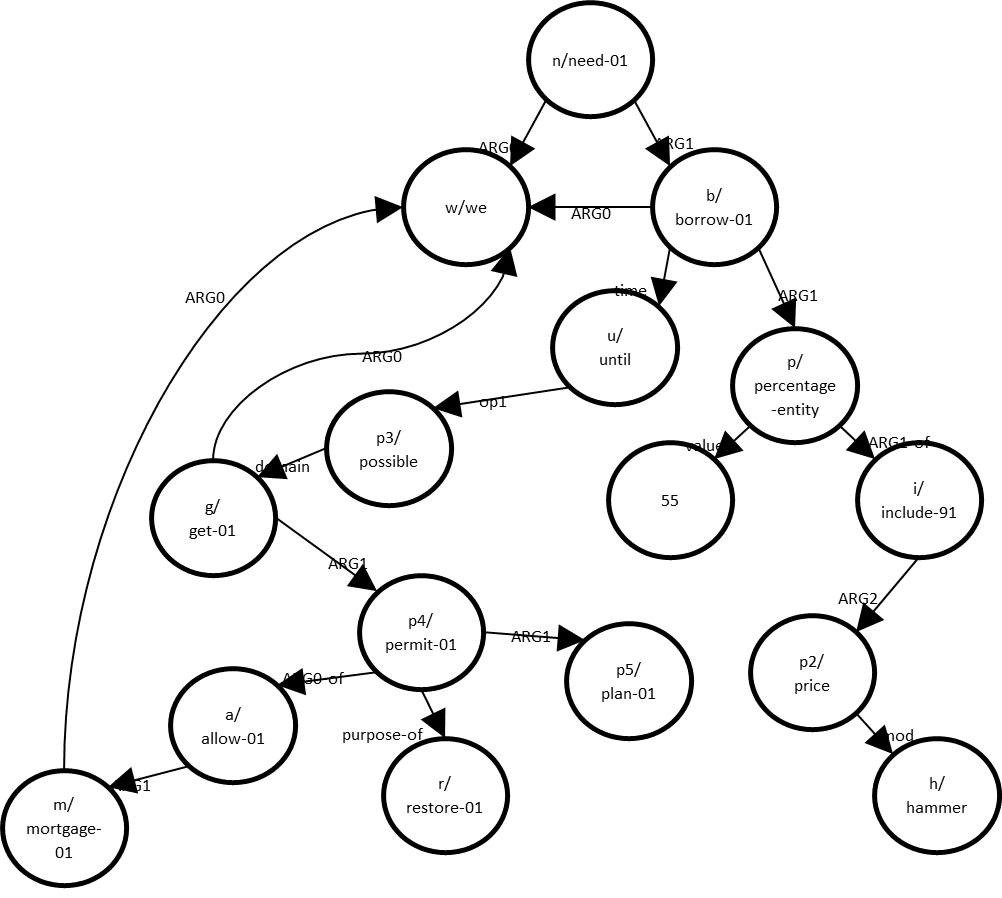}
  \caption{Example Sentence as graph (AMR)}
  \label{fig:example-sentence-graph-AMR}
\end{figure}

Let’s take the sentence “We need to borrow 55\% of the hammer price until we can get planning permission for restoration which will allow us to get a mortgage.” (taken from ~\cite{amrtutorial}). In the textual form of AMR, this can be parsed into Fig. \ref{fig:example-sentence-AMR}. The graphical form of this structure can be found in Fig. \ref{fig:example-sentence-graph-AMR}.

\begin{figure}[h]
  \includegraphics[width=\columnwidth]{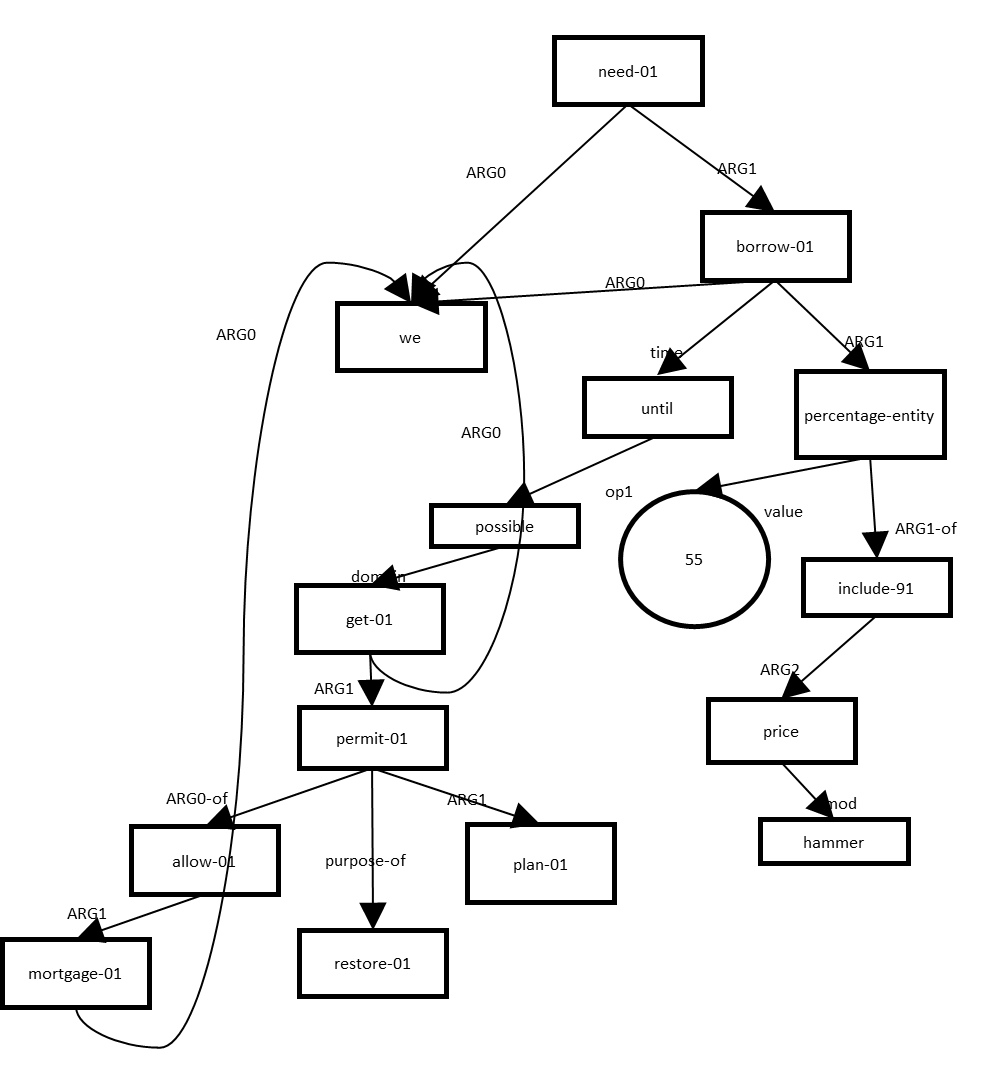}
  \caption{Example Sentence as \textsuperscript{Meta}SRL++}
  \label{fig:example-AMR-as-MetaSRL++ }
\end{figure}

Given the method mentioned above, a corresponding \textsuperscript{Meta}SRL++ Semantic Graph looks quite similar (see Fig. \ref{fig:example-AMR-as-MetaSRL++ }).

\subsection{Uniform Meaning Representation}
\label{s:umr}

\begin{figure}[h]
  \includegraphics[trim={0 0.5cm 0 0},clip,width=\columnwidth]{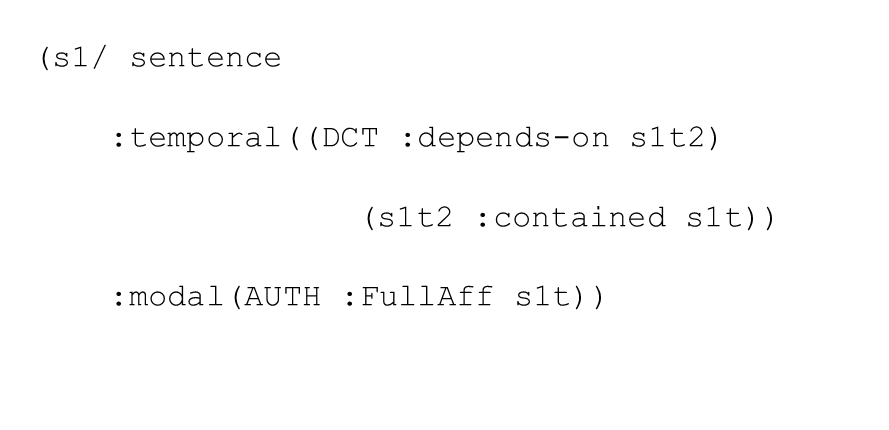}
  \caption{Example UMR Snippet}
  \label{fig:umr6}
\end{figure}

\begin{figure}[h]
  \includegraphics[width=\columnwidth]{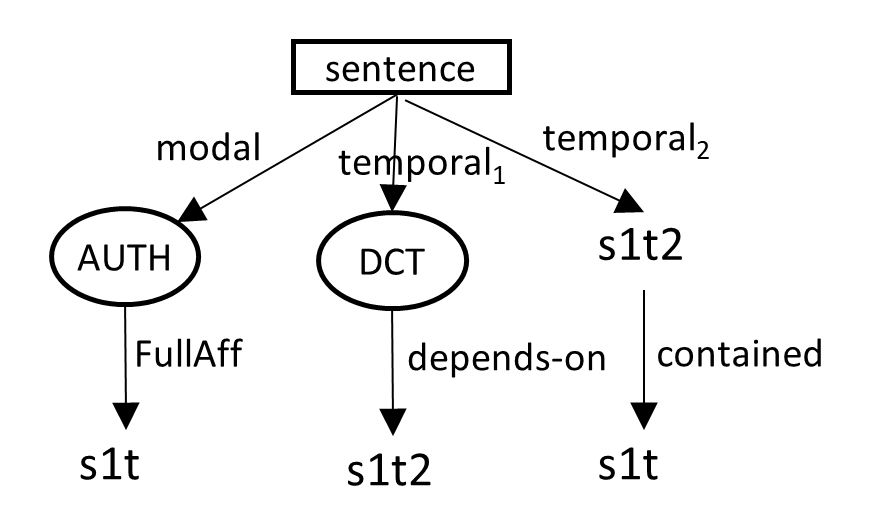}
  \caption{Snipet Conversion}
  \label{fig:umr3}
\end{figure}

\begin{figure}[h]
  \includegraphics[width=\columnwidth]{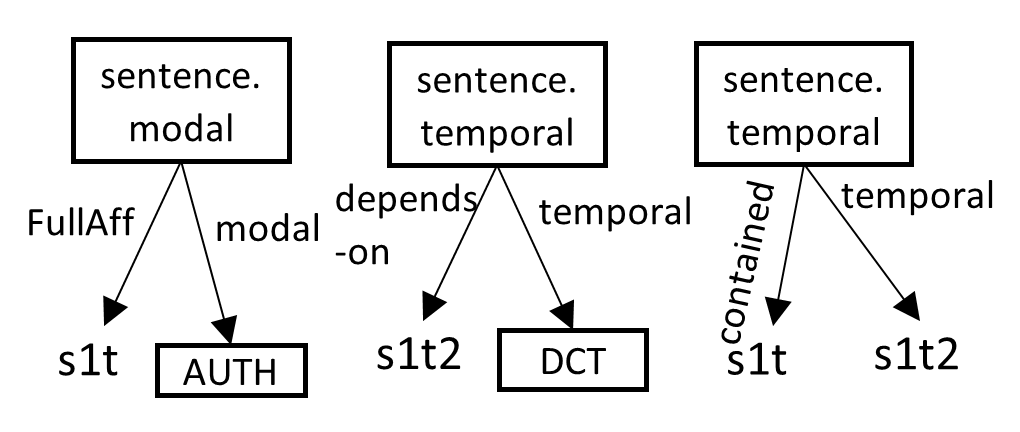}
  \caption{Alternative Conversion}
  \label{fig:umr2}
\end{figure}

Uniform Meaning Representation (UMR) is based on AMR for the (intra) sentence structures and adds semantic document structures like temporal and modal dependencies, and co-reference relations. These add two issues to the way we transformed AMR structures into Semantic Graphs. First, AMR Constants cannot longer be replaced by Entities so easily as dependencies require also Constants to have outgoing edges (which would be forbidden by modelling them as \textsuperscript{Meta}SRL++ Entities). This is not a big problem, we can also model them as concepts. The second issue is that the additional UMR relations cannot be modelled as Roles any longer. Consider the UMR structure in Fig. \ref{fig:umr6} (cited after ~\cite{umrguideline}) and their AMR-like conversion into Semantic Graphs in Fig. \ref{fig:umr3}. The s1t2 reference is the value of the temporal role of the sentence concept. The s1t2 elements has also a contained role to s1t. If we now replace s1t2 by an edge from sentence to s1t2, the relation between the incoming temporal and the outgoing contained role is broken. As a solution to this problem, we incorporate the sentence concept into their immediate roles, model them as concepts and add the transitive edges to these concepts (see Fig. \ref{fig:umr2}).

\subsection{Universal Conceptual Cognitive Annotation}
\label{s:ucca}

The main goal of the Universal Conceptual Cognitive Annotation (UCCA) is to graph-visualise and annotate natural languages using just semantic categories. Only semantic categories are actively annotated, while distributional regularities are learned implicitly by a statistical parser. The graph's representation of semantic differentiation is its primary concern rather than distributional regularities. The collection of relations and their arguments makes up the UCCA semantic representation. The relationships that each layer represents are specified. Each layer specifies the relations which he represents. The foundational layer is designed to cover the entire text so that each word is in at least one node. The nodes of the graphs are called “units”. A unit may be either: 
\begin{enumerate}
\item A terminal or
\item Several elements that are jointly viewed as a single entity. 
\end{enumerate}
A Non-terminal unit will be composed of a single relation and its arguments or it may contain secondary relations as well. 
The UCCA graph follows three main rules: (i) Each unit is a node, (ii) Descendants of non-terminal units are the sub-units, (iii) Non-terminal nodes “only represent the fact that their descendants form a unit so they do not bear any features”~\cite{abend-rappoport-2013-ucca}. In UCCA, the foundational layer views the text as a collection of ``Scenes'', which describes ``some movement or action, or a temporally persistent state'' and ``one main relation, which is the anchor of the Scene'' ~\cite{abend-rappoport-2013-ucca}.

\begin{figure}[h!]
\centering
\includegraphics[width=\columnwidth]{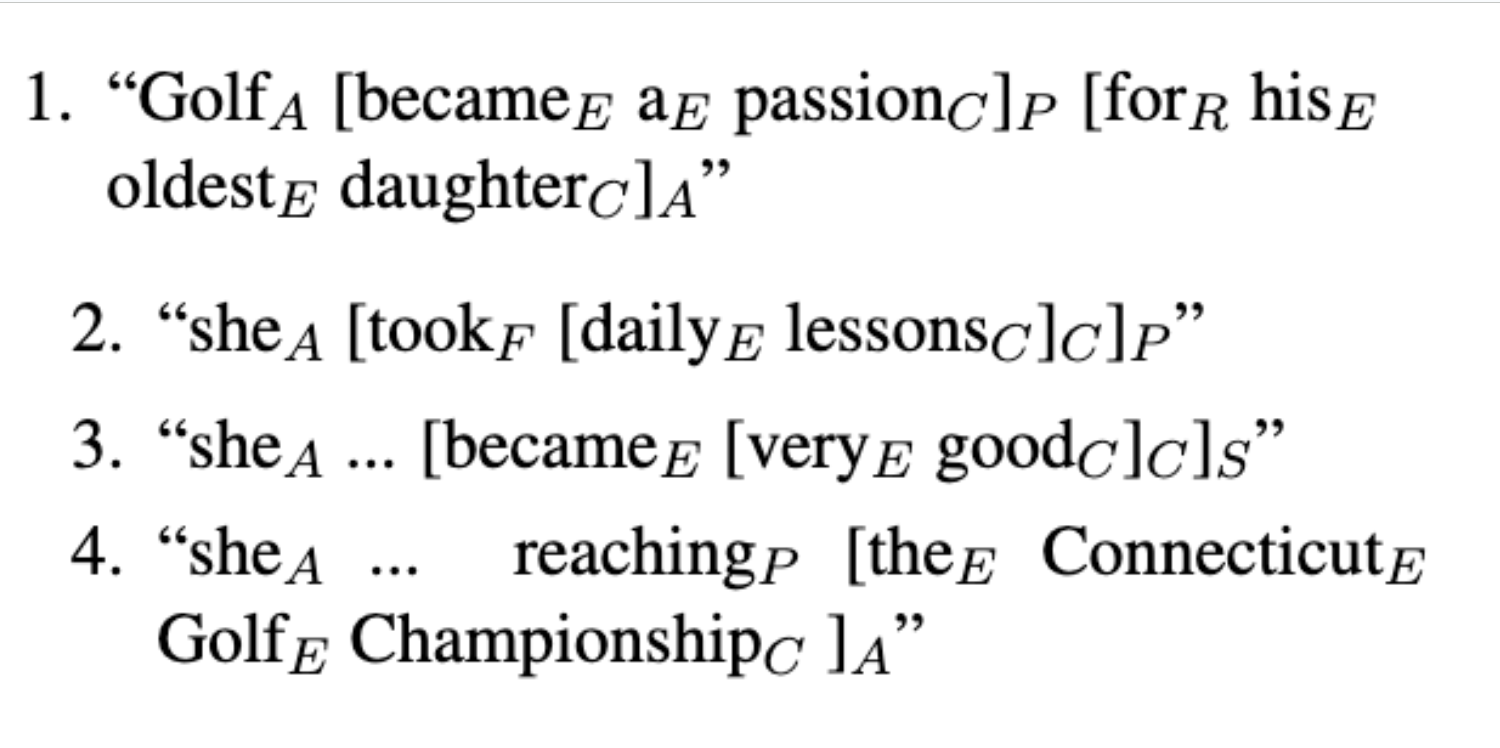}
\caption{An example of UCCA for the sentence ``Golf became a passion for his oldest daughter: she took daily lessons and became very good, reaching the Connecticut Golf Championship''. Taken from \cite{abend-rappoport-2013-ucca}}
\label{fig:ucca}
\end{figure}

\subsubsection{From UCCA to \textsuperscript{Meta}SRL++ }

From a structural point of view, UCCA graphs consist of unlabelled non-terminal nodes, of labelled edges and of terminal nodes that consist of smaller text units (e.g. words).

These graphs can be converted to \textsuperscript{Meta}SRL++ by replacing:

\begin{itemize}
    \item unlabelled non-terminal nodes by a single concept (e.g. UCCA.Unit) that is always the same
    \item labelled edges by labelled edges
    \item terminal nodes by Entities of a suited class with the text unit as a value
\end{itemize}

\begin{figure}[h!]
\centering
\includegraphics[width=\columnwidth]{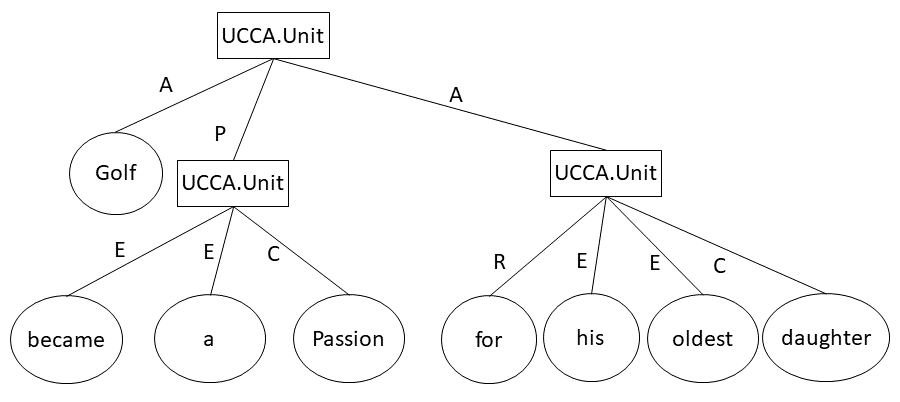}
\caption{Semantic Graph converted from UCCA Graph}
\label{fig:srlucca}
\end{figure}

Using this recipe, e.g. the first sentence of Fig. \ref{fig:ucca} can be converted to the Semantic Graph in Fig. \ref{fig:srlucca}. 

\section{Conclusion \& Further Work}

We argued that one reason for the lack of resources for deep semantic analysis is the lack of a common, uniform representation scheme for deeper semantics that is able to represent all aspects of semantics. We further discussed the idea that the reason for this lack of such a representation scheme is the lack of a common modelling scheme that allows to model semantics. We presented \textsuperscript{Meta}SRL++, our proposal for a uniform modeling scheme for all types of semantic information. We demonstrated how our modelling scheme can be used to convert two heterogeneous semantic data examples into a common format that can be used to export and import semantic data. We explained related work and what the novelty of our approach compared to these approaches is.

In the future, we plan to extend \textsuperscript{Meta}SRL++ to SRL++, a semantic representation scheme based on \textsuperscript{Meta}SRL++. To that end we foresee a way to define concepts and to establish an infrastructure to browse and edit existing concept definitions and contribute new ones. We think that a separation between basic and composed concepts (the latter consisting of basic and other composed concepts) will allow for an efficient usage of SRL++-encoded semantics by applications. Finally, we want to examine approaches to create a minimal set of basic concepts that offers a viable basis for covering a large semantic space. We hope that this will serve as a step towards a larger number of semantic resources and tools, as well as a step towards better neural semantic representations.



\bibliographystyle{acl_natbib}
\bibliography{references}


\end{document}